\documentclass{article}%
\usepackage[T1]{fontenc}%
\usepackage[utf8]{inputenc}%
\usepackage{lmodern}%
\usepackage{textcomp}%
\usepackage{lastpage}%
\usepackage{array}
\usepackage{geometry}%
\usepackage{color,soul}
\usepackage{multicol}%
\usepackage{graphicx}%
\usepackage{lipsum}%
\usepackage{caption}%
\usepackage{booktabs}%
\usepackage{ragged2e}%
\usepackage{breqn}  
\title{DPDR: A novel machine learning method for the Decision Process for Dimensionality Reduction}%
\author{Jean{-}Sébastien Dessureault, Daniel Massicotte}%
\date{\today}%
\begin{document}%
\normalsize%
\maketitle%
\justify%
\section*{ABSTRACT}%
\label{sec:ABSTRACT}%
This paper discusses the critical decision process of extracting or selecting the features in a supervised learning context. It is often confusing to find a suitable method to reduce dimensionality. There are pros and cons to deciding between a feature selection and feature extraction according to the data's nature and the user's preferences. Indeed, the user may want to emphasize the results toward integrity or interpretability and a specific data resolution. This paper proposes a new method to choose the best dimensionality reduction method in a supervised learning context. It also helps to drop or reconstruct the features until a target resolution is reached. This target resolution can be user-defined, or it can be automatically defined by the method. The method applies a regression or a classification, evaluates the results, and gives a diagnosis about the best dimensionality reduction process in this specific supervised learning context. The main algorithms used are the Random Forest algorithms (RF), the Principal Component Analysis (PCA) algorithm, and the multilayer perceptron (MLP) neural network algorithm. Six use cases are presented, and every one is based on some well-known technique to generate synthetic data. This research discusses each choice that can be made in the process, aiming to clarify the issues about the entire decision process of selecting or extracting the features.\newline

\noindent Feature extraction; feature selection; Random Forest algorithm; PCA algorithm; MLP neural network

\begin{multicols}{2}%

\section{Introduction}%
\label{sec:Introduction}%

When we think about reducing dimensionality, there are different schools of thought. It is possible to extract, select, or leave the totality of the features intact. It is not always clear for the data scientist when it is time to choose the best method and parameters. Several points must be taken into account, and too often, this decision is taken empirically with some tries and errors.  
This research proposes a new method that combines supervised and unsupervised machine learning algorithms resulting in a better dataset reduction of dimensionality. It has input, output, and metrics to make the right choice. This decision between feature extraction and feature selection leads to, for instance, a regression or a classification since the method applies to a supervised learning context. It is based on the machine learning methods RF, PCA, and MLP.

This method can help solve several types of problems. Choosing between a feature selection and a feature extraction can sometimes be obvious, but it is not always the case. Furthermore, it is most of the time arduous to know how many features to remove to downsize to a correct level of data resolution. This method can support the visualizing process of data in 2D or 3D graphics or even in more dimensions in a radar graphic. It can help to reduce the overfitting problem, to optimize the accuracy and simplify a model, and speed up the processing time. In summary, this method can help solve Bellman's "curse of dimensionality \cite{bellman_dynamic_1957}, choosing the proper method between feature selection and feature extraction, and selecting the correct number of features to downsize. 

This novel method is inspired by a similar method named DPDRC \cite{dessureault_dpdrc_2022}. The principle is roughly the same, but this first method can only be applied in an unsupervised learning context. This DPDRC method must then generate some labels (a clustering consistency named "Silhouette index") to evaluate the importance of the features. The Feature Ranked Silhouette Decomposition (FRSD) algorithm \cite{yu_ensemble_2020} is used to do so.  It solves this evaluation of features for clustering using a Silhouette Index (SI) metric. It generates SI for every possible combination of features. It loops for each value of \textit{k} (the number of clusters) in a k-means clustering algorithm. An RF algorithm replaces the FRSD method to evaluate the feature selection in this new method. PCA is used in both methods for feature extraction. DPDRC finishes the process with k-means clustering since the context is unsupervised. In this new method, an MLP neural network is used to process regression or classification according to the nature of the data. Hence, DPDRC applies in an unsupervised learning context (for clustering), and DPDR applies in a supervised learning context (for regression and classification). Both are useful to make the right reduction of dimensionality choice. 

To evaluate the importance of the features in a supervised learning context, we have to first use a RF algorithm \cite{biau2016random} \cite{gulea_how_2019} \cite{paul_improved_2018}, based on decision trees and Bayesian network \cite{ronaghan_mathematics_2019}. There are myriads of applications. In \cite{chang_network_2017}, there is an example of an RF application that evaluates the feature importance in detecting a network intrusion. The method needs to lower the cardinality of the dataset to process a Support Vector Machine (SVM), and the RF helps to find the best features to keep in the dataset. 

Principal Component Analysis (PCA) is a practical algorithm to extract features and reduce the dimensionality of a dataset \cite{noauthor_spectral_nodate}. It consists of linear transformations to convert a set of correlated variables into a set of linearly uncorrelated variables. 

Others like \cite{chen_city_2008} use a PCA algorithm combined with cluster analysis (CA) to study social-economic indexes (i.e., non-agriculture population; gross industry output value; the business volume of post and telecommunications; local governments revenue, and others.) The analysis is applied to 17 counties and cities. In this example, a PCA algorithm retrieves the first and second principal components (PC1 and PC2). According to PC1 and PC2, the CA classifies the cities into four classes of growth poles. Researches like \cite{noauthor_big_nodate}\cite{noauthor_lessons_2015} also use a PCA algorithm to extract features in big data and smart cities.  

An MLP is also executed in this method. Although this method is an ancient one based on the works of Rosenblatt \cite{rosenblatt_perceptron_1958}, it is still a basic but useful kind of neural network. A recent book 
\cite{taud_multilayer_2018} gives a new look at today's usages of this method. Using an MLP, \cite{bounds_multilayer_1988} presents a method for diagnosing low back pain and sciatica. A chapter in \cite{park_chapter_2016} is about the MLP usage for ecological models. Finally, \cite{kwon_parallel_2017} needs MLP to process magnetic resonance (MR) images and improve the signal quality. 

The 6 datasets used in the research comes from the \textit{scikit learn} framework \cite{avila_scikit-learn_2017} \cite{kramer_scikit-learn_2016}. The functions \textit{sklearn.datasets.make_classification()} \cite{noauthor_sklearndatasetsmake_classification_nodate} and \textit{sklearn.datasets.make_regression()} \cite{noauthor_sklearndatasetsmake_regression_nodate} allow to perfectly generate some different datasets for classification and regression, respectively, according to some desired parameters.  Each function can generate a specific number of features and rows, according to a certain distribution.  Both functions are commonly used in the scientific litterature. For instance, the function \textit{sklearn.datasets.make_regression()} is used in \cite{holt2021training} and \textit{sklearn.datasets.make_classification()} is part of the process in \cite{lemaitre2017imbalanced}. 
This first presents a method that allows a model to "forget" a user data from a machine learning system when this user is removed. This is for security purposes.  
The second presents \textit{imbalanced-learn}, an open-source python toolbox aiming at providing a wide range of methods to cope with the problem of imbalanced datasets frequently encountered in machine learning and pattern recognition.

The main contribution of this paper is a new machine learning method to automatize the reduction of the dimensionality decision process (feature extraction or feature selection) according to the data scientist preferences (target resolution of data, interpretability oriented or integrity oriented parameters). It also helps to determine the amplitude of the reduction according to the user preference or according to an equation. 

The next sections of this paper are organized with the following structure: Section \ref{sec:Methodology} describes the proposed methodology. Section \ref{sec:Results} presents the results. Section \ref{sec:Discussions} discusses the results and their meaning and Section \ref{sec:Conclusion} concludes this research.

\section{Methodology}%
\label{sec:Methodology}%

\subsection{Datasets and features}%
\label{subsec:Sub_methodology_datasets_features}%
The 6 datasets have been generated to create regression problems and classification problems. The number of samples and the number of features vary from dataset to dataset and are documented in table \ref{table:table_datasets}.  The data about regression problems have been generated with \textit{sklearn.datasets.make_regression()} function.  The classification problems are generated with \textit{sklearn.datasets.make_classification()}.  The features are all composed with \textit{float} type values.  For the regression problems, a target field (a \textit{float}) defines the regression value. For the classification problems, a class field (an \textit{integer}) defines the category of the sample. \newline

\begin{minipage}[htb]{1.0\columnwidth}
\captionof{table}{Datasets list}%
\begin{tabular}{llll} \hline 
\label{table:table_datasets}%
\textbf{Dataset} & \textbf{Type} & \textbf{Rows} & \textbf{Feat.}  \\
\hline
1. & Regression & 500 & 5  \\
2. & Classification & 500 & 5  \\
3. & Regression & 500 & 25  \\
4. & Classification & 500 & 25  \\
5. & Regression & 1000000 & 8  \\
6. & Classification & 10000 & 100  \\

\hline
\end{tabular}
\end{minipage}\newline

For all those datasets, the parameter "random_state" has been set to 1.  It means that always the same data sequence will be generated, leading to perfect reproducibility.  Setting this parameter to "None" would mean a different data sequence each time.

\subsection{Proposed model design}%
\label{subsec:Sub_methodology_model_design}%

The whole proposed model is a combination of several parts using different machine learning algorithms. The result of the process is parameterizable by the data scientist. The first parameter is the nature of the problem (regression or classification). The two following parameters are the preference of the data scientist, whether he prefers results based on interpretability or integrity. For instance, the interpretability parameter could have a value of 40\%, while the integrity parameter could have a value of 60\%, meaning that a bias toward data integrity will be induced. Those two parameters are in a $\alpha$ and $1 - \alpha$ pattern. Those parameters will help orient the choice between a feature selection or a feature extraction method. The last parameter is the required minimum resolution of data. This parameter will help decide the correct number of features to keep in the resulting dataset. If no parameter is sent to the algorithm, it will assume that the integrity and interpretability parameters will have a 50\% value. The resolution parameter will be automatically calculated to obtain the best possible value according to an equation defined later. 

Fig. \ref{fig:architecture} presents the architecture of the complete methodology. There are two pipelines in the model: 1. The feature selection pipeline, and 2. the feature extraction pipeline. It consists of processing both of them. At the end of each one, a score will be delivered. This score will be used to make the right choice between a feature selection and a feature extraction and, ultimately, to return the reduced dataset to the data scientist. 

Let us start with the first pipeline: The feature selection pipeline. 
This part uses an RF algorithm to evaluate the importance of the features, sorting the features in descending order of importance. This sorted list is then filtered to keep only the first most important feature, just enough to reach the required minimum resolution parameter. The less important ones are dropped. The reduced ordered features list is then sent to the input layer of an MLP. This MLP performs a regression or a classification according to the parameter and the used dataset. The MLP returns an accuracy score for this feature selection pipeline. 

Let us see the second pipeline: The feature extraction pipeline. Since this part is about feature extraction, the importance of the features is analyzed by a PCA algorithm. Like in the feature selection pipeline, the features are placed in descending order of importance. The method counts the  \textit{n} number of features required to reach the value of the minimum resolution parameter. Instead of simply dropping the features, all the features are extracted to generate  \textit{n} new features called "Principal Component 1", "Principal Component 2", ...,  "Principal Component n",  (PC1, PC2, ... PCn).   The resulting reduced list feeds the MLP neural network that performs a regression or a classification. In the end, the MLP returns an accuracy score. This is the output of the extraction pipeline.  

Ultimately, the two resulting scores feed two equations that compute the interpretability and integrity scores. The higher score will determine the best method between feature selection (interpretability) and feature extraction (integrity). 

\noindent \includegraphics[width=\columnwidth]{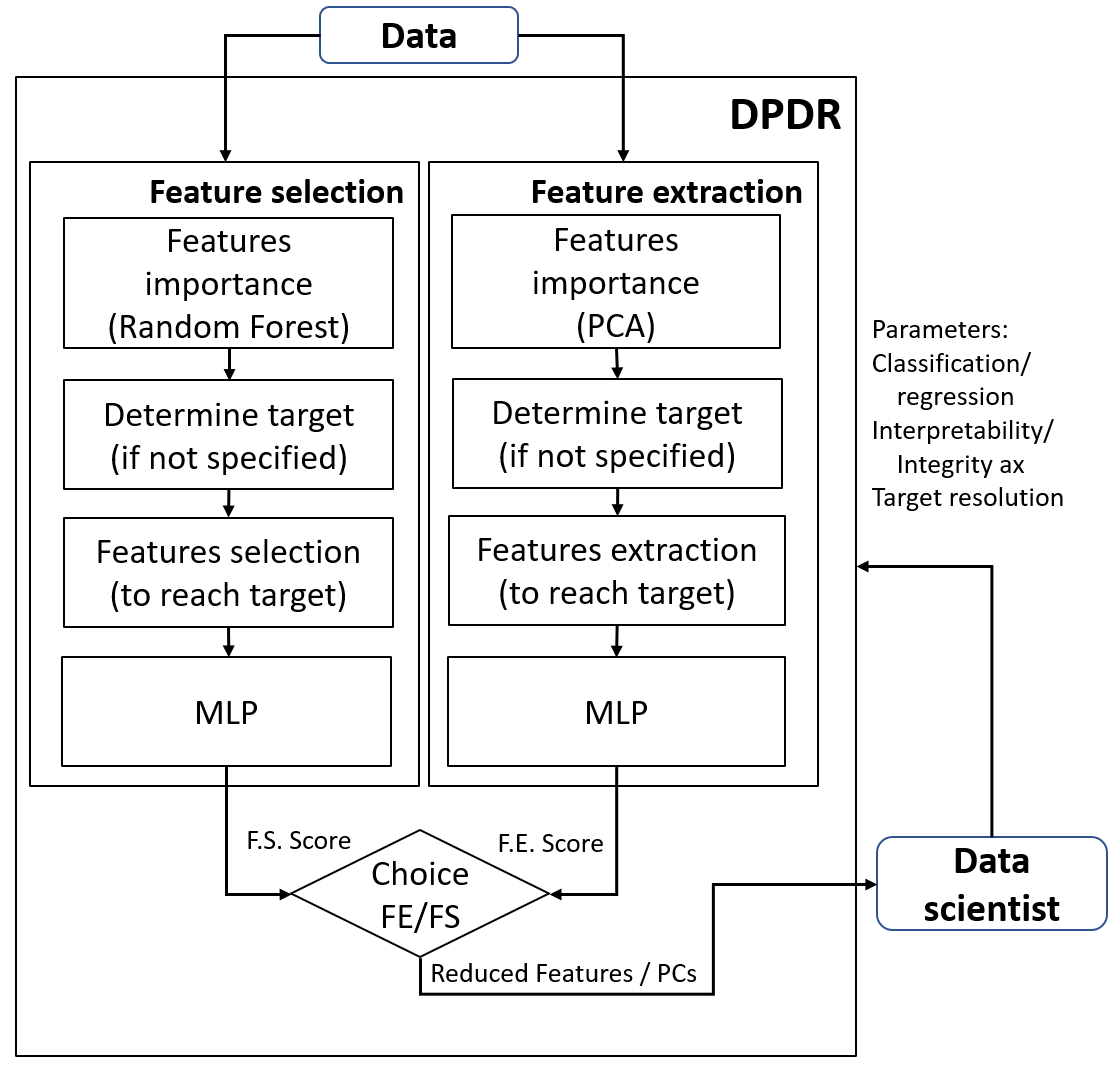}
\captionof{figure}{Architecture of the methodology \newline \label{fig:architecture}}%
The following section describes each part of the process and its machine learning algorithms. \newline

\subsection{Evaluation of features using Random Forest}%
\label{subsec:Sub_methodology_RF}%

There are several machine learning methods to analyze the importance of the features in a dataset. One effective method is the RF algorithm. Being a particular case of a Bayesian network, an RF is based on multiple decision trees. A single decision tree may lead to some bias. Having tenths of decision trees gives better results, as it pools all the outcomes to return the most frequent answer. The RF algorithms are popular for solving some regression problems and some classification problems. Furthermore, the solution can be visualized in a tree graph where the nodes represent the decision, the branches represent the possible outcomes, and the leaves represent the possible answers (combined with their probability). 

Having the ability to compute a whole Bayesian probability tree, the RF algorithm is also capable of calculating the feature importance in the decision process. It can evaluate the role of each feature in producing the outcome. If a feature plays an essential role in the final decision, the value will be high. On the opposite, the less a feature is important in the final decision, the lower the importance value will be. All the feature importances are normalized, and the sum of all the features is 1. The final result, a list of features and their importance, is needed at the next step in the feature selection.

\subsection{Feature selection using Random Forest}%
\label{subsec:Sub_methodology_RF}%

This part of the whole method selects only the most critical features, just enough to reach the minimum resolution parameter value. For instance, if a required minimum resolution is set to 80\%, and there are five features of importance 35\%, 30\%, 15\%, 10\%, and 10\%. Then the first three features would be selected. The sum of the first three more important features gives 80\%. Hence two more features are not required. 
The resulting reduced list of features will be helpful to the neural network in the next part. 

\subsection{Evaluation of features using PCA}%
\label{subsec:Sub_methodology_PCA_evaluation}%
The algorithm executes a PCA algorithm on the dataset. At this time, the number of extracted features equals the number of features. The result is simply the importance of the features in an unsupervised learning context. The target (for the regression problem) or the class (for the classification problem) is not considered. One of the outputs of this process is the explained variance between the features. After dividing this explained variable for each feature by the total amount of variance, it gives the importance of each feature for the extraction process. 

This list of features and their importance will be essential in the next step of the process. 

\subsection{Feature extraction using PCA}%
\label{subsec:Sub_methodology_PCA_extraction}%
At this level, having the importance of the feature, the goal is to extract the data of the feature to downsize to the number of features required to reach the minimum resolution parameter. The newly created features are named "principal components" (PCs). The first principal component (PC1) has the highest possible variance compared to the other principal components. The second principal component (PC2) has the second-highest possible variance, and so on. The advantage of this method is that the data of every feature has been extracted and taken into account. The disadvantage is that all the feature names are lost and replaced by non-significative names like PC1, PC2, and PCn.  

The resulting list of PCs will be the input of the MLP neural network in the next part.

\subsection{Determining target resolution}%
\label{subsec:Sub_Selecting_resolution}%
This part can be found in both feature selection and feature extraction pipelines.  In the case there is no specified value for the target_resolution parameter, the system will propose a good one, based on some basic rules. \newline

Before explaining the rules and the equations, let us define what could be a "good resolution". When a data scientist intuitively tries to find a "good place" to divide (to keep on the left and drop on the right) the features in descending importance order, he looks at several points: 1. He makes a sum of the most important features to ensure he has enough data resolution. 2. He looks at the differences between the importance of the features that are not included in the minimum resolution needed. 3. He selects the most important difference in this range of features and cuts there. The key point is: having a maximum resolution with a minimum dimension. This intuitive method is the base of this proposed algorithm. \newline

The following are the formal rules to implement this concept. First, every feature that is part of the last 30\% of resolution and has an importance of less than 3\% will not be kept. After having applied this rule, equations \eqref{eq:resolution1}, \eqref{eq:resolution2}, \eqref{eq:resolution3} and \eqref{eq:resolution4} are defining each step of the algorithm.  \eqref{eq:resolutionFinal} resumes whole equation system.  All the features are in descending order of importance and multiplied by 10. 

\begin{equation}%
\lambda_{f} = \sum_{i=1}^{f.nb.} \phi_{i}  
 \label{eq:resolution1}%
\end{equation}

Where $\phi$ is the normalized importance of the feature, \textit{f.nb.} is the number of features, $\lambda$ is the resolution of the data at feature \textit{f}. 

\begin{equation}%
\Delta_{f} = \phi_{f} - \phi_{f+1}  
 \label{eq:resolution2}%
\end{equation}

Where $\Delta_{f}$ is, at feature \textit{f}, the difference of importance between one feature and the next feature.

\begin{equation}%
w\Delta_{f} = \left( \Delta_{f} \right)^{2}    
 \label{eq:resolution3}%
\end{equation}

Where $w\Delta_{f}$ is the weighted difference of the features at feature \textit{f}. 

\begin{equation}%
\alpha = Max_{i=1}^{n-1} \lambda_{f} + w\Delta_{f} 
 \label{eq:resolution4}%
\end{equation}

Where $\alpha$ is the best score.

\begin{equation}%
\alpha = \max_{i=1}^{n-1}\left( \sum_{i=1}^{f} \phi_{i} + \left( \phi_{f} - \phi_{f+1} \right)^{2}   \right) 
 \label{eq:resolutionFinal}%
\end{equation}

Fig. \ref{fig:equation_explained_1} explained some crucial components of the equations using a typical graphic of the feature's importance.  This representation is when f = 2 in the iteration  .  

\noindent \includegraphics[width=\columnwidth]{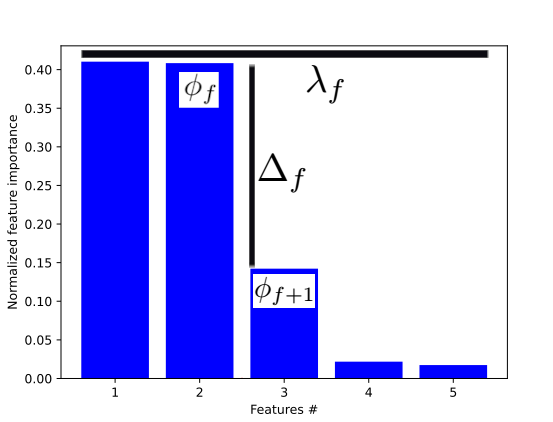}
\captionof{figure}{Equations variables explained, for f=2.\newline \label{fig:equation_explained_1}}%

Fig. \ref{fig:equation_explained_2} shows the graph of the score for each feature/PC. It can be seen that every value has been multiplied by 10. The blue part is the sum of the features' importance until the feature itself. The orange part is the importance difference between the feature itself and the following feature. The sum of those two values is the score. The higher score is kept as the $\alpha$ value, representing where the cut is made.

\noindent \includegraphics[width=\columnwidth]{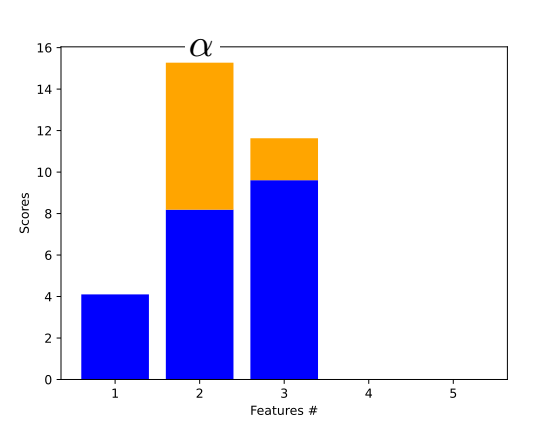}
\captionof{figure}{Representation of the scores determining the best place to cut.\newline \label{fig:equation_explained_2}}%

Fig. \ref{fig:equation_explained_3} shows the final graph at the end of the process.  The vertical red line is situated where the cut is made for feature selection or feature extraction.     

\noindent \includegraphics[width=\columnwidth]{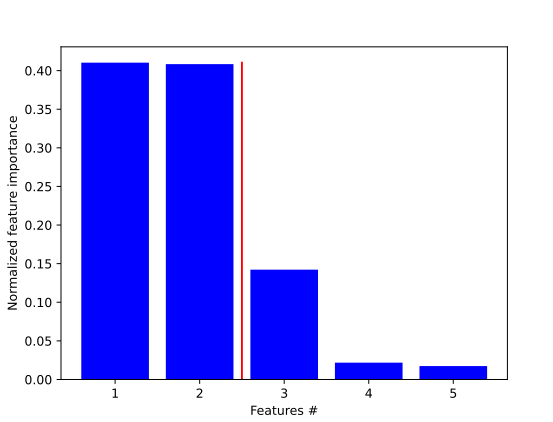}
\captionof{figure}{Final graph at the end of the process.\newline \label{fig:equation_explained_3}}%

\subsection{Regression and classification using neural network}%
\label{subsec:Sub_methodology_MLP}%

Both pipelines (feature selection and feature extraction) end with a neural network of type MLP. The reason is to try the two new reduced sets of data, the first from the feature selection process and the second from the feature extraction process. According to the nature of the data and parameter, MLP executes a regression or classification on both datasets. The algorithm uses k-Fold cross-validation (where k=10) to ensure that the gradient descent does not overfit the result. The algorithm usually quits before the maximum epoch parameter allowed.

The architecture configuration of the MLP is shown in the table \ref{table:table_MLP_archi}.  \newline

\begin{minipage}[htb]{1.0\columnwidth}
\captionof{table}{MLP architecture}%
\begin{tabular}{ll} \hline 
\label{table:table_MLP_archi}%
\textbf{Parameters} & \textbf{Values}  \\
\hline
Solver & Stoch. Gradient Descent \\ 
Learning rate & 0.001 \\
TOL & 0.0001 \\
Cross-validation & K-Fold, k=10 \\
Input neurons & (as many as the number \\
 & of features/PCs) \\
Hidden layers & 1 \\
Neurons in hidden layer & 1.2 X (\# input neurons)\\
Output neurons & (as many as the number \\
& of targets/classes)\\
Max Epoch & 100000\\
\hline
\end{tabular}
\end{minipage}\newline 

The result is the accuracy percentage for feature selection and feature extraction pipeline. The two decision equations will need those results in the next step.  

\subsection{Choice between feature extraction and feature selection}%
\label{subsec:Sub_choice_FE_FS}
First, let us define the important "interpretability/integrity" axis. It is crucial since it is at the heart of the choice between a feature selection and a feature extraction. The feature selection favours interpretability. It has the advantage of keeping each feature's name intact, except for those that are dropped. For instance, if a dataset is composed of 4 features named \textit{Age}, \textit{Weight}, \textit{Height}, and \textit{Blood pressure}, and the feature selection process drops the \textit{Height} feature, it will remain \textit{Age}, \textit{Weight}, and \textit{Blood pressure}. This feature selection process favours interpretability. Although, some features and the contained information are excluded from the reduced dataset, disadvantaging the data integrity. 
On the other part, feature extraction favours the integrity of the data. When downsizing a dataset, the information included in all the features is considered. Better integrity of data is kept compared to feature selection. The counterpart is that all of the names of the features are lost, being replaced by abstract other names called "Principal component", or PC. Using the same example as above, our features \textit{Age}, \textit{Weight}, \textit{Height}, and \textit{Blood pressure} would become \textit{PC1}, \textit{PC2} and \textit{PC3} after the feature extraction process. A feature called \textit{PC1}, is less interpretable than a feature named \textit{Age}, for instance. 

Two parameters are defining this interpretability/integrity axis: 1.  \textit{interpretability-oriented} and  2. \textit{integrity-oriented}. Both are on a scale of 0\% to 100\%, representing the importance of the axe. The sum of those percentages must be equal to 100\%. Both parameters describe how important this value is. 

The decision can be taken after the result of both the feature selection pipeline and the feature extraction pipeline. The decision considers both the performance of the MLP regression/classification and the preference of the data scientist for interpretability or integrity. 

An \textit{interpretability score} is computed as defined in (\ref{eq:interpretability_score})

\begin{equation}%
interpret_{s} = interpret_{p} * MLP.accuracy_{fs} 
 \label{eq:interpretability_score}%
\end{equation}\newline

\textit{interpret_{s}} is the interpretability score, \textit{interpret_{p}} is the interpretability parameter, and  \textit{MLP.accuracy_{fs}} is the accuracy of the MLP algorithm after a feature selection. Similarly, an \textit{integrity score} is computed in (\ref{eq:integrity_score})

\begin{equation}%
integ_{s} = integ_{p} * MLP.accuracy_{fe} 
 \label{eq:integrity_score}%
\end{equation}\newline

\textit{integ_{s}} is the integrity score, \textit{integ_{p}} is the integrity parameter, and  \textit{MLP.accuracy_{fe}} is the accuracy of the MLP algorithm after a feature extraction.

The two scores are compared, and the higher will be the algorithm's choice. Then, the new set of features/PCs will be returned according to the algorithm choice (feature selection or feature extraction). The justification of the decision is also displayed and returned to the data scientist. \newline
The justification of the decision is composed of 1. Normalized importance of features after the RF process 2. Normalized importance of features after the PCA process 3. Best MLP Accuracy for Feature selection 4. Best MLP Accuracy for feature extraction 5. Interpretability score 6. Integrity score 7. Chosen method (selection of extraction) 8. Number of selected features to obtain the target resolution (if feature selection is used) 9. Number of principal components to obtain the target resolution (if feature extraction is used) 10. Resolution of data reached. \newline

\section{Results}%
\label{sec:Results}%

\subsection{MLP accuracy before reduction of dimensionality}%
\label{subsec:dataset}%
An MLP (regressor or classifier, according to the nature of the datasets) algorithm has been executed on every dataset to establish an accuracy reference. Table \ref{table:table_MLP_reference} shows the accuracy of the MLP model before having reduced dimensionality on datasets.\newline

\begin{minipage}[htb]{1.0\columnwidth}
\captionof{table}{MLP accuracy for datasets before reduction of dimensionality}%
\begin{tabular}{llllr} \hline 
\label{table:table_MLP_reference}%
\textbf{Dataset} & \textbf{Type} & \textbf{Rows} & \textbf{Feat.} & \textbf{Acc.}  \\
\hline
1. & Regression & 500 & 5 & 99.99 \\
2. & Classification & 500 & 5 & 93.20 \\
3. & Regression & 500 & 25 & 99.99 \\
4. & Classification & 500 & 25 & 96.20 \\
5. & Regression & 1000000 & 8 & 99.99 \\
6. & Classification & 10000 & 100 & 98.86 \\
\hline
\end{tabular}
\end{minipage}\newline

\subsection{Cases using different parameters}%
\label{subsec:SubResultsScenarios}%

The following part describes six cases.  The cases are based on the dataset presented in table \ref{table:table_datasets} and use different parameter values to test the proposed method.
Each case shows how the orientation and the resolution parameters affect the decision process. \newline

\noindent \textbf{Case 1: Simple selection} \newline
This scenario is based on dataset 1, which is a regression problem. We suppose that it is equally important to orient the results toward data interpretability and data integrity and that a good feature resolution is needed.  The value of \textit{interpretability-oriented} = 0.5, \textit{integrity-oriented} = 0.5 and \textit{target-resolution} = 90\% has been used. Table \ref{table:Scenario1} shows the results using this configuration. \newline

\begin{minipage}[htb]{1.0\columnwidth}
\centering
\captionof{table}{Algorithm results using \textit{interpretability_oriented} = 0.5, \textit{integrity_oriented} = 0.5 and \textit{target_resolution} = 90\%}%
\label{table:Scenario1}
\begin{tabular}{|l|r|}
\hline 
\textbf{Metrics} & \textbf{Values} \\ 
\hline 
  Dataset							& 1 \\
  Problem type					 	& Regression  \\
  Initial nb. of features			& 5 \\
  Nb. of rows						& 500 \\
  MLP Accuracy (Selection)			& 0.961  \\  
  MLP Accuracy (Extraction)			& 0.941  \\
  Interpretability score 	  		& 0.481  \\
  Integrity score 		  			& 0.471  \\
  Chosen method 		  			& SELECTION  \\
  Number of selected features/PCs 	&  \\  
  to obtain target resolution 		& 4  \\
  Resolution (target)	  			& 90.0\%  \\
  Resolution (actual)	  			& 96.4\%  \\
\hline 
\end{tabular}
\end{minipage}\newline

In this scenario, the parameter \textit{interpretability-oriented} has an equal value to the \textit{integrity-oriented} value. When (\ref{eq:interpretability_score}) and (\ref{eq:integrity_score}) are calculated, the interpretability score is higher than the integrity score because of the higher score of the MLP accuracy. To reach 90\% of the resolution, we must use the best 4 of the five features. It reaches a resolution of 96.4\%.   Fig. \ref{fig:Bar1} shows the result of the dimensionality reduction after the feature selection process. 

\noindent \includegraphics[width=\columnwidth]{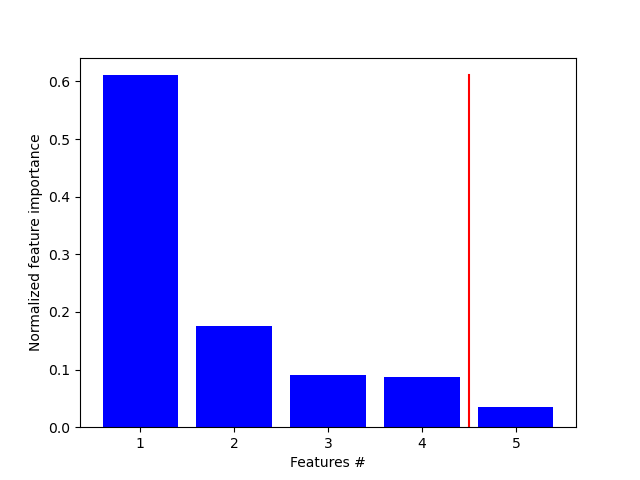}
\captionof{figure}{Result of the reduction of dimensionality for case 1. \newline \label{fig:Bar1}}%

The X-axis contains the features, and the Y-axis is the feature importance level. Features are placed in descending order. The Red line shows the division where the features are kept (left side) and dropped (right side). 

The conclusion is that if the interpretability and integrity parameters are equal, the decision will be taken as the result of the MLP accuracy based on the data. In other words, if the data scientist does not have a preference, then the algorithm will go for a better match in the data. \newline

\noindent \textbf{Case 2: Simple extraction} \newline
Relying on dataset 2, this scenario addresses a classification problem.  The value of \textit{interpretability-oriented} = 0.4, \textit{integrity-oriented} = 0.6 and \textit{target-resolution} = 75\% has been used. Table \ref{table:Scenario2} shows the results with this configuration. \newline

\begin{minipage}[htb]{1.0\columnwidth}
\centering
\captionof{table}{Algorithm results using \textit{interpretability_oriented} = 0.4, \textit{integrity_oriented} = 0.6 and \textit{target_resolution} = 75\%}%
\label{table:Scenario2}
\begin{tabular}{|l|r|}
\hline 
\textbf{Metrics} & \textbf{Values} \\ 
\hline 
  Dataset							& 2 \\
  Problem type					 	& Classification  \\
  Initial nb. of features			& 5 \\
  Nb. of rows						& 500 \\
  MLP Accuracy (Selection)			& 0.936  \\  
  MLP Accuracy (Extraction)			& 0.934  \\
  Interpretability score 	  		& 0.374  \\
  Integrity score 		  			& 0.560  \\
  Chosen method 		  			& EXTRACTION  \\
  Number of selected features/PCs 	&  \\  
  to obtain target resolution 		& 4  \\
  Resolution (target)	  			& 75.0\%  \\
  Resolution (actual)	  			& 87.3\%  \\
\hline 
\end{tabular}
\end{minipage}\newline

Fig. \ref{fig:Bar2} presents the result of the dimensionality reduction for this second case.

\noindent \includegraphics[width=\columnwidth]{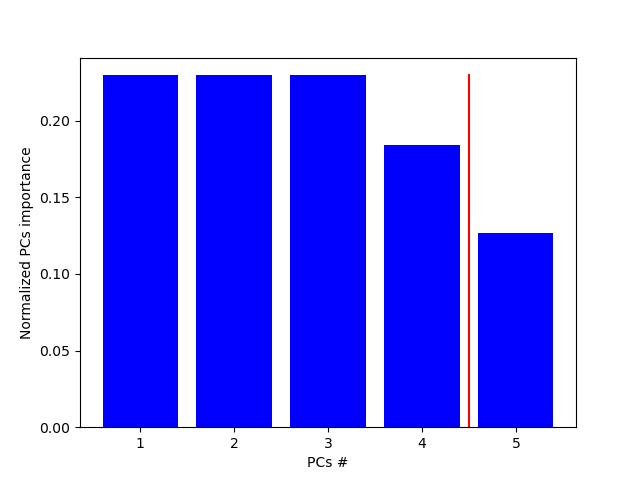}
\captionof{figure}{Result of the reduction of dimensionality for case 2. \newline \label{fig:Bar2}}%

In this scenario, the parameter \textit{interpretability-oriented} has a slightly lower value than the \textit{integrity-oriented value}. Calculating (\ref{eq:interpretability_score}) and (\ref{eq:integrity_score}), the interpretability score is  lower than the integrity score, so the feature extraction process is preferred.  To reach 75\% of the resolution, we must reduce it to 4 features, and it reaches a resolution of 87.3\%.    

Hence, if the accuracy results are about the same for both selection and extraction, the interpretability and integrity parameters will be the tiebreaker.  \newline

\noindent \textbf{Case 3: 25 features, no predetermined resolution parameter} \newline
This scenario addresses a regression problem. The number of features (25) is higher than in the two first scenarios (5). The parameters orient the results toward integrity, and the parameters do not give the resolution of the feature. The value of \textit{interpretability-oriented} = 0.8, and \textit{integrity-oriented} = 0.2 has been used. Table \ref{table:Scenario3} displays the results with this configuration. \newline

\begin{minipage}[htb]{1.0\columnwidth}
\centering
\captionof{table}{Algorithm results using \textit{interpretability_oriented} = 0.8, \textit{integrity_oriented} = 0.2 and \textit{target_resolution} = N.A.}%
\label{table:Scenario3}
\begin{tabular}{|l|r|}
\hline 
\textbf{Metrics} & \textbf{Values} \\ 
\hline 
  Dataset							& 3 \\
  Problem type					 	& Regression  \\
  Initial nb. of features			& 25 \\
  Nb. of rows						& 500 \\
  MLP Accuracy (Selection)			& 0.963  \\  
  MLP Accuracy (Extraction)			& 0.912  \\
  Interpretability score 	  		& 0.777  \\
  Integrity score 		  			& 0.183  \\
  Chosen method 		  			& SELECTION  \\
  Number of selected features/PCs 	&  \\  
  to obtain target resolution 		& 6  \\
  Resolution (actual)	  			& 82\%  \\
\hline 
\end{tabular}
\end{minipage}\newline

The result of the feature selection is presented in Fig. \ref{fig:Bar3}.

\noindent \includegraphics[width=\columnwidth]{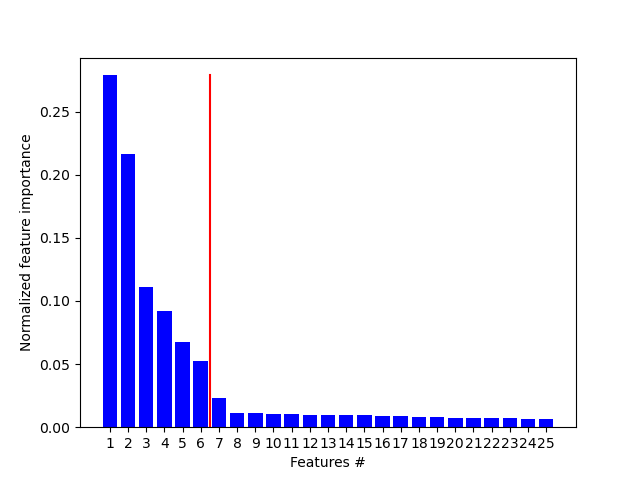}
\captionof{figure}{Result of the reduction of dimensionality for case 3. \newline \label{fig:Bar3}}%

The feature selection is chosen without surprise since the interpretability parameter value is significantly higher than the integrity parameter.   The main point for this case is that, for 25 parameters, the algorithm has chosen a reasonable resolution of its own, according to the equations presented in \ref{subsec:Sub_Selecting_resolution}.  

We can conclude that the feature selection has been made correctly and automatically.  \newline

\noindent \textbf{Case 4: 25 features, no parameters at all} \newline
Using dataset 4, this case addresses a classification problem.  The number of features is still 25, as in the last scenario. In this case, no parameter are passed to the algorithm so the default values are used.  The default value of \textit{interpretability-oriented} is 0.5, and \textit{integrity-oriented} is 0.5.  The \textit{target-resolution} is automatically calculated to fit the equations in \ref{subsec:Sub_Selecting_resolution}.  Table \ref{table:Scenario4} shows the results with this configuration. \newline

\begin{minipage}[htb]{1.0\columnwidth}
\centering
\captionof{table}{Algorithm results using no parameter}%
\label{table:Scenario4}
\begin{tabular}{|l|r|}
\hline 
\textbf{Metrics} & \textbf{Values} \\ 
\hline 
  Dataset							& 4 \\
  Problem type					 	& Classification  \\
  Initial nb. of features			& 25 \\
  Nb. of rows						& 500 \\
  MLP Accuracy (Selection)			& 0.946  \\  
  MLP Accuracy (Extraction)			& 0.946  \\
  Interpretability score 	  		& 0.473  \\
  Integrity score 		  			& 0.473  \\
  Chosen method 		  			& SELECTION  \\
  Number of selected features/PCs 	&  \\  
  to obtain target resolution 		& 1  \\
  Resolution (actual)	  			& 83\%  \\
\hline 
\end{tabular}
\end{minipage}\newline

The result of the feature selection is presented in Fig. \ref{fig:Bar4}.

\noindent \includegraphics[width=\columnwidth]{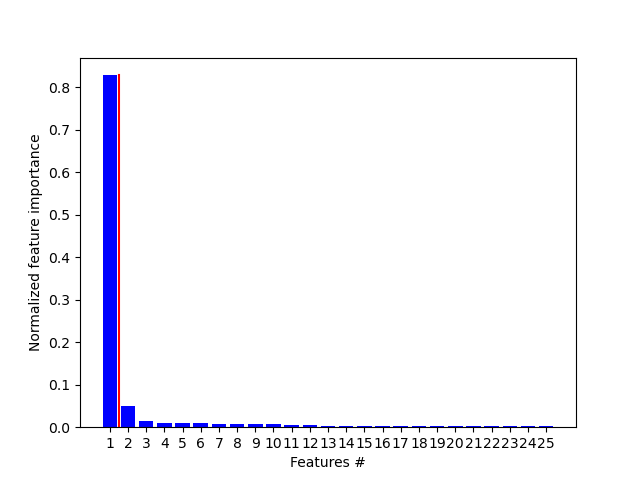}
\captionof{figure}{Result of the reduction of dimensionality for case 4. \newline \label{fig:Bar4}}%

In this scenario, the data scientist's preference is unknown since no parameter is passed to the algorithm. The MLP accuracy is the same for feature selection and feature extraction. In this rare case, interpretability is preferred because the feature's labels are kept. There is no specified parameter for the data resolution, so the algorithm calculates it. Only one feature was required, and a resolution of 83\% was reached.   

The algorithm can deal without any parameter if the data scientist does not want to interfere. Also, only one feature can be selected if it has enough importance. \newline  

\noindent \textbf{Case 5: 8 features, 1,000 000 rows and no predetermined resolution parameter} \newline
A classification problem is executed using this dataset 5. The model's scalability is tested with only eight features but 1,000 000 rows. There is more weight to the integrity (0.6) rather than interpretability (0.4). No resolution parameter is specified. Table \ref{table:Scenario5} shows the results. \newline

\begin{minipage}[htb]{1.0\columnwidth}
\centering
\captionof{table}{Algorithm results using \textit{interpretability_oriented} = 0.4, \textit{integrity_oriented} = 0.6 and \textit{target_resolution} = None}%

\label{table:Scenario5}
\begin{tabular}{|l|r|}
\hline 
\textbf{Metrics} & \textbf{Values} \\ 
\hline 
  Dataset							& 5 \\
  Problem type					 	& Regression  \\
  Initial nb. of features			& 8 \\
  Nb. of rows						& 1000000 \\
  MLP Accuracy (Selection)			& 0.980  \\  
  MLP Accuracy (Extraction)			& 0.993  \\
  Interpretability score 	  		& 0.392  \\
  Integrity score 		  			& 0.600  \\
  Chosen method 		  			& EXTRACTION  \\
  Number of selected features/PCs 	&  \\  
  to obtain target resolution 		& 7  \\
  Resolution (actual)	  			& 98\%  \\
\hline 
\end{tabular}
\end{minipage}\newline

The result of the feature selection is presented in Fig. \ref{fig:Bar5}.
\noindent \includegraphics[width=\columnwidth]{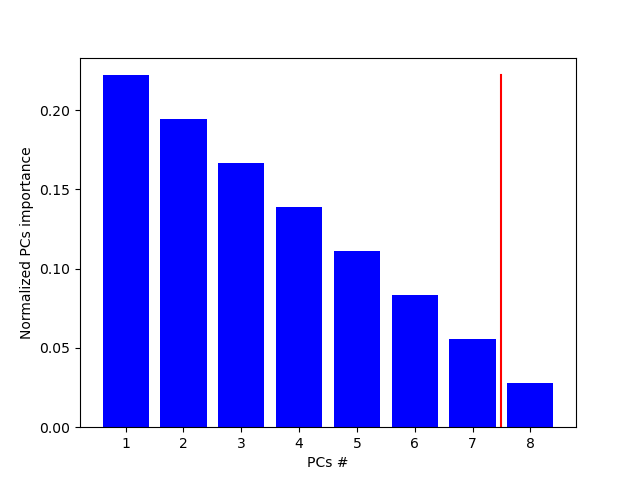}
\captionof{figure}{Result of the reduction of dimensionality for case 5. \label{fig:Bar5}}%

This scenario aims to prove that the algorithm still can perform using more data. There are a million rows in this dataset. Note that to reach the calculated 98\% of the resolution, we must extract seven features of the eight original features.  

The conclusion is that the method is still working, with a million data. The scalability aspect of this method will be discussed in \ref{subsec:scalability}.   \newline
\noindent \textbf{Case 6: 100 features, 10 000 rows and no predetermined resolution parameter} \newline
Using dataset 6, this scenario addresses a classification problem.  The number of features is high to 100 but has fewer rows (10 000) than in the last scenario. The parameters orient the results toward interpretability (0.8).  Table \ref{table:Scenario6} displays the results with this configuration. \newline

\begin{minipage}[htb]{1.0\columnwidth}
\centering
\captionof{table}{Algorithm results using \textit{interpretability_oriented} = 0.8, \textit{integrity_oriented} = 0.2 and \textit{target_resolution} = None}%
\label{table:Scenario6}
\begin{tabular}{|l|r|}
\hline 
\textbf{Metrics} & \textbf{Values} \\ 
\hline 
  Dataset							& 6 \\
  Problem type					 	& Classification  \\
  Initial nb. of features			& 100 \\
  Nb. of rows						& 10000 \\
  MLP Accuracy (Selection)			& 0.875  \\  
  MLP Accuracy (Extraction)			& 0.988  \\
  Interpretability score 	  		& 0.700  \\
  Integrity score 		  			& 0.198  \\
  Chosen method 		  			& SELECTION  \\
  Number of selected features/PCs 	&  \\  
  to obtain target resolution 		& 1  \\
  Resolution (actual)	  			& 57\%  \\
\hline 
\end{tabular}
\end{minipage}\newline

At the opposite of the other cases, no figure is presented for this case since the 100 features make the graphic unreadable.\newline 

This method is still efficient, having 100 features. In this case, only one feature represents 57\% of the resolution. Therefore, the selection was made on only one feature. This case aims to prove the scalability of the method. It will be discussed later in \ref{subsec:scalability}. 

\subsection{Scalability}%
\label{subsec:scalability}%
The proposed algorithm was tested on a server with the following configuration: 11th Gen Intel(R) Core(TM) i9-11900K @ 3.50GHz with 64Gb of RAM and an operating system Windows 11 64 bits. Having the execution time for each case, this part shows the algorithm's scalability according to the number of features and the number of data (rows in the dataset).

Fig. \ref{fig:Scalability1} and \ref{fig:Scalability2} help to validate the scalability of the proposed method. They show the training time with two different large-scaled configurations. X axe represents the number of data trained, and Y represents the required time to train this data. Blue points are the mean of the evaluation points, and the blued area is the range of the evaluation points.

First, let us examine case number 5. In this scenario, a million rows were used to train the model. Fig.\ref{fig:Scalability1} shows the execution time for each evolution of training data through training time using eight features in an MLP after a feature extraction process.

\centering
\noindent \includegraphics[width=\columnwidth]{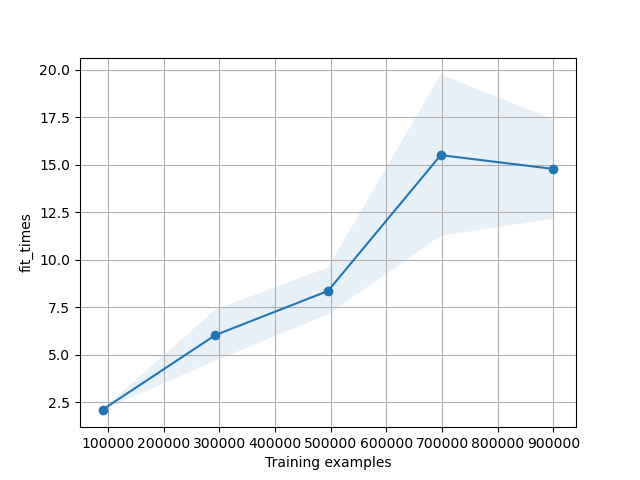}
\captionof{figure}{Scalability after feature extraction (case 5) using a MLP\newline \label{fig:Scalability1}}
\justifying 

According to the nature of the reduced data, the line is not perfectly linear, but a trend can be seen. We can conclude that the method does not take too much time, even when processing a million of data and eight features. At least, we can see that the growth is not exponential.

Let us examine case number 6, where only 10000 rows are used to train the model, but with 100 features. Fig.\ref{fig:Scalability2} shows the training time using 100 features in an MLP after a feature selection process.

\centering
\noindent \includegraphics[width=\columnwidth]{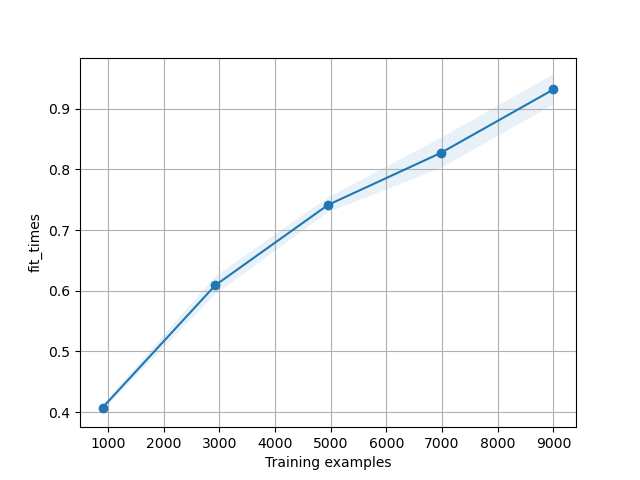}
\captionof{figure}{Scalability after feature selection (case 6) using MLP\newline \label{fig:Scalability2}}
\justifying 

This line is quite linear, too, showing that there is no scalability problem with using 100 features. No exponential growth is observed. \newline

In Fig. \ref{fig:Scalability3} and Fig.\ref{fig:Scalability4}, 11 new datasets has been generated using the same methodology as described in \ref{subsec:Sub_methodology_datasets_features}. Both graphics show the total execution time of the whole methodology (including the evaluation of the features, the feature selection, the feature extraction, the MLP, and the decision process) 

In Fig.\ref{fig:Scalability3} the processing time is according to the number of features shown. 

\centering
\noindent \includegraphics[width=\columnwidth]{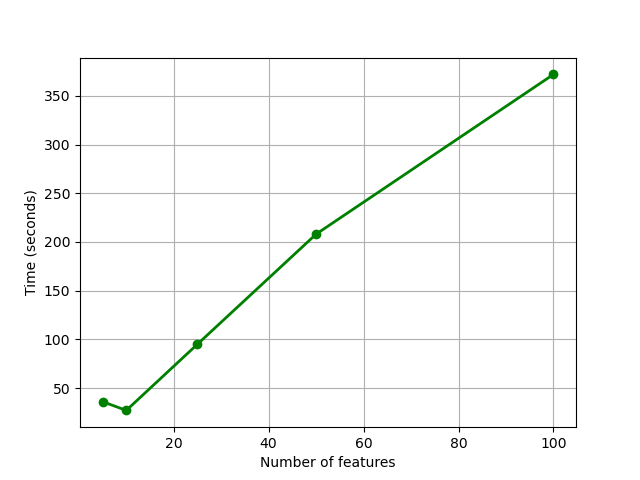}
\captionof{figure}{Running time for different number of  features\newline\label{fig:Scalability3}}
\justifying 

We can see that the line is almost perfectly linear, meaning that the method is scalable up to 100 features.   Case 5 shows that the results are correct.  Fig. \ref{fig:Scalability4} shows the processing time according to the number of data (rows).  \newline
\centering
\noindent \includegraphics[width=\columnwidth]{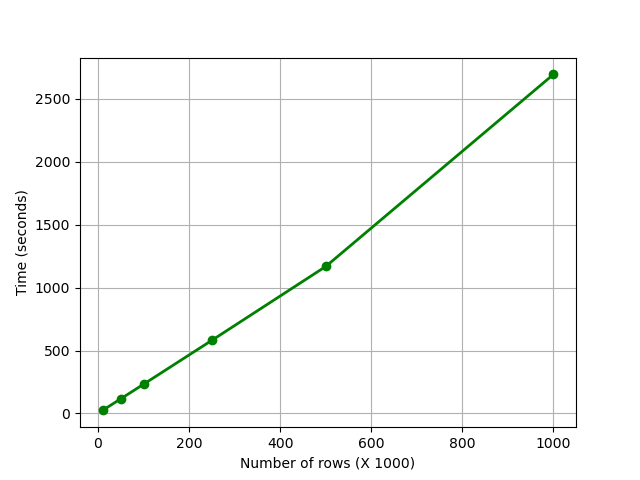}
\captionof{figure}{Running time for different number of rows\newline\label{fig:Scalability4}}
\justifying 

Here again, the line is almost linear.  It means that this method is scalable to a million rows.  Case 6 shows the correctness of the results.

\subsection{Method validation}%
\label{subsec:SubResultsValidation}%

This last part validates the method. 250 realistic random cases have been generated to verify the classification of each point by the algorithm. The simulation chooses a random accuracy (between 0.6 and 0.99) for feature selection and feature extraction. The feature extraction and selection correlate in a +/- 0.2 range. The \textit{interpretability-oriented} parameter is also randomly selected, and the \textit{integrity-oriented} parameter is defined using 1 - (the \textit{interpretability-oriented} value). Using \ref{eq:interpretability_score} and \ref{eq:integrity_score}, the decision is taken whether the point is classified in interpretability (feature selection) or integrity (feature extraction).  Fig. \ref{fig:Validation1} shows the classification of the points.  Red points will use a feature extraction process (for better integrity), and the blue points will use a feature selection process (for better interpretability). A black line divides the interpretability and integrity domains. 

\centering
\noindent \includegraphics[width=\columnwidth]{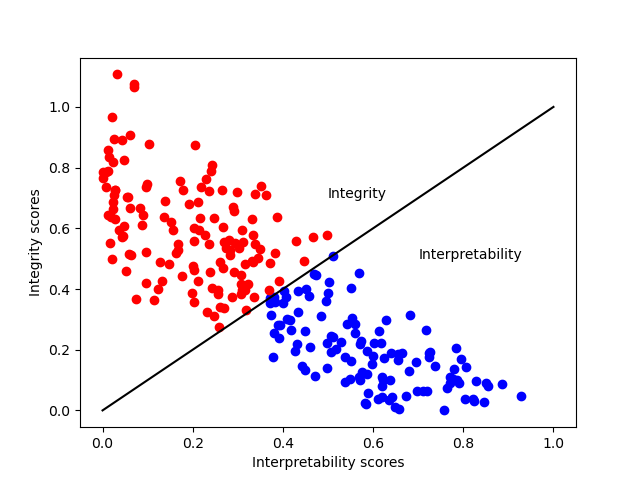}%
\captionof{figure}{Classification of the interpretability scores and integrity scores.\newline \label{fig:Validation1}}
\justifying 

Note that the points will never have a high value on both axes. The interpretability importance parameter is the inverse of the integrity importance parameter ($\alpha$ and $1 - \alpha$). The algorithm never misclassifies the point since it uses a threshold to do the classification. This part of the method is straightforward, but it is a better way to make the best decision having all the data gathered in the previous parts of the whole algorithm.  \newline

\section{Discussions}%
\label{sec:Discussions}%
This new method can be compared to a method called "DPDRC, Decision Process for Dimensionality Reduction before Clustering" \cite{dessureault_dpdrc_2022}. Both have a similar utility, although having a critical difference. DPDRC is proposed in an unsupervised learning context (namely, the clustering process), and DPDR is proposed in a supervised learning context (for regression and classification). Both methods aim for the same goal: To assist a data scientist in the decision process of reduction of dimensionality. An essential addition of DPDR is that it can determine its degree of resolution according to its equation. DPDRC uses the FRSD \cite{yu_ensemble_2020} algorithm to evaluate the feature importance in an unsupervised context, generating a silhouette index for cluster consistency metric. It also ends by generating some clusters using the k-means algorithm. Since dimensionality reduction is also helpful in a supervised learning context, the proposed DPDR method replaces the FRSD algorithm with the RF algorithm. The label is already included in the datasets, so there is no need to generate new ones as FRSD does in DPDRC. At the end of the process, the clustering is replaced by regression or classification, depending on the dataset type. Both are complementary and have their utility, according to the supervised or the unsupervised context. 

Results of the compared methods are shown in \ref{table:comp_DPDRC_DPDR}. Cases come from both Case 1 of the respective study. The number from each case cannot be compared directly since they use two different datasets. DPDRC is an unsupervised clustering problem, and DPDR is a supervised regression problem. Although, the table shows some typical results and magnifies the differences between these two approaches.  \newline

\noindent\begin{minipage}[htb]{1.0\columnwidth}
\centering
\captionof{table}{Comparison between DPDRC and DPDR metrics}%
\label{table:comp_DPDRC_DPDR}
\begin{tabular}{|l|r|r|}
\hline 
\textbf{Metrics} & \textbf{DPDRC} & \textbf{DPDR}  \\ 
\hline 
  Context & Unsupervised & Supervised \\
  Problem type & Clustering & Regression \\ 
  Best FS silhouette index & 0.390 & N.A. \\
  Best FE silhouette index & 0.353 & N.A. \\
  Best FS MLP accuracy & N.A. & 0.961 \\
  Best FE MLP accuracy & N.A. & 0.941 \\
  Interpretability score 	  & 0.351 &  0.481 \\
  Integrity score 		  & 0.035 &  0.471 \\
  Chosen method 		  & Selection & Selection  \\
  Number of selected features & & \\  
  to obtain target resolution & 7 & 4 \\
  Resolution 			  & 88.3\% & 94.4\% \\
  Best number of clusters (k) & 3 & N.A.  \\
\hline 
\end{tabular}
\end{minipage}\newline

This table shows that DPDR follows the same DPDRC concept but in a supervised learning context. Those methods complement the typical feature selection and feature extraction processes.  

Comparing those two methods (DPDR and DPDRC) with RF (for a feature selection) and PCA (for feature extraction), we note that those last methods do not take any decision regarding the best point to cut. On the opposite, DPDR and DPDRC are built to orient the process toward RF or PCA and keep the right number of features according to a predetermined or calculated target resolution. 

This method can help in many circumstances. There are several situations where choosing a good reduction of dimensionality technique is not clear, and this method is helpful. Also, this method helps to know how many features (representing how much resolution) to extract or select.   

Example 1: There are many situations where reducing dimensionality can help visualize the data. Data are especially easy to visualize in 2D or 3D, but the dataset could have higher cardinality (6, for instance). In this case, there is a need to downsize the number of features to better visualize the data. The same can apply with an even higher cardinality (let's say 30). By reducing the cardinality of the dataset to 6, it would then be possible to present the data with a radar graphic. What would be the best method to reduce dimensionality in that situation? There is no clear answer to that. Data can be more interpretable if every original feature label is kept. But it is also good to have data with better integrity, even if the labels are as abstract as "PC1", "PC2", etc. So both are good choices, and the data scientist just has to enter his preferences in the algorithm parameters.  

Example 2: The dimensionality reduction can help reduce the overfitting problem. Dimensionality reduction finds a lower number of variables or removes the least important features from a dataset. The model complexity is reduced, and often some noise vanishes in this process, reducing the risk of overfitting. Both feature selection and feature extraction are helping to do so. The works of Richard Bellman \cite{bellman_dynamic_1957} explain all the disadvantages of having too much dimension in a dataset. Bellman calls it "The curse of dimensionality".  

Other examples: There are several other situations where the decision is not obvious, though it is required to downsize the dataset. This Bellman's "Curse of dimensionality" has other disadvantages. The dimensionality reduction reduces the execution time for training and testing the model. It simplifies the model and improves its accuracy. This new method not only helps make a good decision about the method of reduction of dimensionality, but it also helps to find how many features or PCs to keep and discard. 


The main contribution of this paper is to define a new complete method in a supervised learning context that makes the right choice of dimensionality reduction. It also helps to choose how many features to remove to downsize to the needed data resolution. This process is done according to the data scientist's preferences. 

Two algorithms were needed to evaluate the feature's importance: RF and PCA. The first evaluates the feature importance for feature selection and the second for feature extraction. RF returns the importance of the feature according to the feature's utility to find the target feature or its class, and PCA gives the importance of the features according to the covariance with other features. Both explain the importance of the features in the dataset, using their definition of "importance". 

The decision process uses two basic equations, (\ref{eq:interpretability_score}) and (\ref{eq:integrity_score}), based on the data scientist's preferences concerning interpretability and integrity. If not specified, the resolution to keep is specified by (\ref{eq:resolutionFinal}). The process ends by performing regression or classification according to the nature of the dataset. The performance is evaluated, and the result of this MLP neural network is returned to the data scientist, supplying a full solution using the correctly reduced-sized dataset.  

Fig. \ref{fig:Scalability1}, \ref{fig:Scalability2}, \ref{fig:Scalability3} and \ref{fig:Scalability4} are showing, for the tested values, that there are no scalability problem. Regarding the scalability, we can conclude that there is no problem using up to a million data (rows) and 100 features. The figure shows certain linearity between the quantity of data (rows and features) and the execution time. No exponential curve is observed.  
 
The validation of the method is shown in Fig. \ref{fig:Validation1}. It shows that the algorithm makes a good decision of feature selection or extraction using 250 generated data and parameters.

\section{Conclusion}%
\label{sec:Conclusion}%

In a supervised learning context, this paper proposed a novel method to select the correct dimensionality reduction technique between a feature selection and a feature extraction and calculate the correct number of features or PCs. It finally executes a regression or a classification to end resolving the problem using the downsized dataset. \newline

In the future, this method can be improved by optimizing the parameters of each algorithm needed (RF, PCA, and MLP). There are many ways to do some regressions and some classifications. Different other types of neural networks can be tried, having different parameters. Some other algorithms like support vector machines (SVM) can be tested. There are many possible configurations of parameters that can be tried to execute the RF algorithm and the PCA algorithm. It would also be possible to test the scalability on a higher volume of data. It can be on over a million rows and over a hundred features. Furthermore, new data presentations can be added to improve this method. 

\section{Acknowledgement}%
\label{sec:Acknowledgement}%
This work has been supported by the "Cellule d’expertise en robotique et intelligence artificielle" of the Cégep de Trois{-}Rivières.

\bibliographystyle{plain}%
\bibliography{paper}

\end{multicols}%
\end{document}